\PassOptionsToPackage{colorlinks,linkcolor=blue,urlcolor=blue,citecolor=blue}{hyperref}

\documentclass[
  journal=proceedings,
  manuscript=article-type,
  year=2025
]{preprint}

\usepackage{amsmath}
\usepackage[nopatch]{microtype}
\usepackage{booktabs}
\usepackage{orcidlink}
\usepackage{threeparttable}
\usepackage{graphicx}
\usepackage{tabularx}
\usepackage{multirow}

\newcommand{\orcidauthor}[2]{#1~\orcidlink{#2}}

\usepackage{amssymb}

\title{Prompt Engineering for Scale Development in Generative Psychometrics}

\author{\orcidauthor{Lara L. Russell-Lasalandra}{0009-0000-3014-1937}}
\affiliation{University of Virginia, Charlottesville, VA, USA}
\email{llr7cb@virginia.edu}

\author{\orcidauthor{Hudson Golino}{0000-0002-1601-1447}}
\affiliation{University of Virginia, Charlottesville, VA, USA}
\email{hfg9s@virginia.edu}


\begin{document}

\begin{abstract}
This Monte Carlo simulation examines how prompt engineering strategies shape the quality of large language model (LLM)--generated personality assessment items within the AI-GENIE framework for generative psychometrics. Item pools targeting the Big Five traits were generated using multiple prompting designs (zero-shot, few-shot, persona-based, and adaptive), model temperatures, and LLMs, then evaluated and reduced using network psychometric methods. Across all conditions, AI-GENIE reliably improved structural validity following reduction, with the magnitude of its incremental contribution inversely related to the quality of the incoming item pool. Prompt design exerted a substantial influence on both pre- and post-reduction item quality. Adaptive prompting consistently outperformed non-adaptive strategies by sharply reducing semantic redundancy, elevating pre-reduction structural validity, and preserving substantially larger item pool, particularly when paired with newer, higher-capacity models. These gains were robust across temperature settings for most models, indicating that adaptive prompting mitigates common trade-offs between creativity and psychometric coherence. An exception was observed for the GPT-4o model at high temperatures, suggesting model-specific sensitivity to adaptive constraints at elevated stochasticity. Overall, the findings demonstrate that adaptive prompting is the strongest approach in this context, and that its benefits scale with model capability, motivating continued investigation of model--prompt interactions in generative psychometric pipelines.
\end{abstract}

\section{Introduction}

Generative artificial intelligence (AI) is increasingly reshaping foundational practices in psychometrics \citep{garrido2025estimating}. Large language models (LLMs) are now routinely used in psychological scale development to automatically content analyze items \citep{fyffe2024}, create embeddings to be used for dimensionality recovery prior to data collection \citep{garrido2025estimating}, generate entire item pools \citep{gotz2024let,hernandez2023ai,lee2023paradigm}, and execute the post-generation checks [e.g., content review, bias assessment, and revision; \citet{lee2025ai}]. We've even seen best-practice recommendations detailing the responsible use of AI within this precise context \citep[see][]{brickman2025,SALAH2025101698}. What is more, these LLM-authored items have already been shown to be adequately comparable to expert-authored items in some contexts \citep{hommel2022transformer}, and \emph{surpass} them in others \citep{martin2025harnessing}.

These developments reflect a broader methodological shift toward treating language itself as a scalable source of psychological information. That is, language can be analyzed, filtered, and structured even before human response data are collected. This emerging set of practices has led to what has been termed \textbf{Generative Psychometrics} \citep{russell2024generative}, a research paradigm that integrates LLM-based generation with quantitative psychometric evaluation to support faster, more scalable, and more reproducible measurement development. Item content is no longer treated as a fixed human-authored input, but as a probabilistic output that can be generated in large quantities and evaluated algorithmically.

Within this field, a new methodology operationalizing psychometrics and LLMs has been proposed: \textbf{Automatic Item Generation with Network-Integrated Evaluation} (\emph{AI-GENIE}; \citealp{russell2024generative}), which uses LLMs to generate large pools of candidate items and then applies network psychometric methods to evaluate and reduce these pools. Although modern LLMs are capable of producing highly fluent and expert-like text without retraining \citep{brown2020language,carlini-extracting,raffel2023exploring}, psychometric scale development requires more than surface-level plausibility. The central challenge is generating items that accurately reflect the intended construct while maintaining adequate structural validity and minimizing redundancy. AI-GENIE addresses this challenge by offering a scalable pipeline that not only generates items but also (deterministically) evaluates them \emph{in silico} using psychometric approaches, improving the efficiency and feasibility of assessment development.

Despite these advances, a critical methodological question remains largely unexplored in this emerging field: how do different \emph{prompt engineering} strategies influence the quality of item pools entering generative psychometric pipelines? Prompt engineering, i.e., the art of effectively communicating with an advanced model to obtain high-quality, desirable output, has become a central component of harnessing the full potential of AI systems across applications \citep{openai_prompt_engineering_guide,russell2024generative}. More formally, it can be defined as the systematic design and optimization of input prompts to guide LLM responses, ensuring accuracy, relevance, and coherence in the generated output \citep{chen2025}. Prompt engineering is now widely recognized as a key element of LLM output quality \citep{ekin2023prompt,fatawi2024empowering,white2023prompt}. What began as an empirical practice has evolved into a structured research domain, with milestones ranging from early structured inputs to modern techniques such as chain-of-thought and self-consistency prompting, all within the broader framework of \emph{in-context learning} \citep{chen2025}. Accordingly, several resources now provide practical guidance for crafting effective prompts (\citealp{ekin2023prompt}; \citealp{giray2023prompt}; see \citealp{openai_prompt_engineering_guide}).

The relevance of a well-articulated prompt cannot be overstated, as prompt quality plays a central role in the consistency and viability of generated outputs \citep{jin2022good,white2023prompt}. Within the context of psychological assessment, prior work has shown that prompt variation influences generated content \citep{dauphin2025llama,DEWINTER2024112729} and can potentially alter the psychometric properties of resulting items, including aspects of validity \citep{lee2025ai,Marengo2025}. However, existing work in generative psychometrics has largely treated prompting as a fixed or secondary design choice. No study has systematically examined how different prompting strategies affect redundancy, structural validity, and item-reduction outcomes within a psychometrically grounded generative framework.

The present research therefore investigates whether, and to what extent, different prompting strategies improve item generation outcomes within the AI-GENIE framework \citep{russell2024generative}, addressing a key gap in the generative psychometrics literature: how prompting choices affect the quality of both the initial item pool and the final, automatically filtered and selected set of items. Specifically, this study examines whether newer, larger language models benefit more from advanced prompt-engineering methods than earlier models, and how this relationship is modulated by the temperature parameter. Finally, the efficacy of these strategies is evaluated across multiple GPT models to assess how model characteristics interact with different prompting approaches.

\subsection{Prompt Engineering and In-Context Learning}

The ``prompt'' is your interface between your human intent and the model's understanding. Because LLMs generate responses by predicting text conditioned on the prompt, even small changes in phrasing, structure, or contextual framing can meaningfully alter the content, style, and reliability of the output \citep{amatriain2024}. Prompt engineering therefore provides a mechanism for ``eliciting'' and ``steering'' the model towards one's objective, influencing both what the model generates and how consistently it does so across repeated trials.

On its surface, prompt engineering may appear to be simply a valuable technical \emph{skill}, perhaps one that is particularly promising for researchers and educators in higher education \citep{lee2025prompt}. However, as AI has advanced into the black-box, exceptionally large and deep models we know today, prompting has increasingly become more than just a craft learned through trial and error. Instead, it has developed into an emerging research paradigm \citep{schulhoff2025}, with growing interest in systematic strategies that improve performance, reduce hallucinations, increase controllability, and support reproducibility \citep{Huang_2025,liu2024jailbreakingchatgptpromptengineering}. Prompt engineering has even begun to take on the characteristics of a formal profession, as organizations invest in training employees to properly prompt LLM-based systems \citep{joshi2025retraining}.

However, prompt engineering is only part of the equation. The model's ability to demonstrate in-context learning (ICL), or its ability to adapt to new tasks from inputted examples without model-parameter tuning \citep{brown2020language}, plays a central role in the efficacy of prompting strategies. That is, model-prompt interaction effects impact output quality \citep{CHEN2025,han2025evaluating,mehta2025scaling}. For example, when \citet{wei2022chain} introduced the chain-of-thought prompting strategy (i.e., an ICL approach that provides the model with a series of reasoning steps), they found that this prompting technique was particularly effective \emph{only} once models reach sufficient scale.

Recent theoretical work reveals once models exceed a critical capacity, they exhibit qualitatively stronger in-context adaptation, including greater stability, reduced sensitivity to sampling temperature, and the ability to internalize complex, set-level constraints from prompts \citep{mehta2025scaling}. As a result, advanced prompting strategies disproportionately benefit newer, larger models, while older models remain more sensitive to stochasticity and prompt variation. Therefore, generative models will be increasingly responsive to advanced prompt design as model capabilities improve. As such, these empirical comparisons will report the model and prompt interaction rather than per-model performance alone.

\subsection{Prompting Techniques}

The present study evaluates four widely used approaches: (1) zero-shot prompting, using both a \emph{basic} and an \emph{expanded} simulation condition; (2) few-shot prompting; (3) persona-based prompting; and (4) adaptive prompting. Each technique can be understood as a distinct way of structuring the four core components of an effective prompt: clear task instructions, relevant contextual information, specification of the desired output format, and provision of any necessary input data \citep{chen2025,russell2024generative}. The techniques vary systematically in how many of these components they activate and how explicitly they do so.

The most minimal prompt engineering scheme is zero-shot prompting, in which the model receives task instructions alone (e.g., ``Generate items measuring conscientiousness'') with no examples or contextual scaffolding \citep{brown2020language}. This strategy is attractive because of its simplicity and scalability, but it places the full burden of inferring the desired format, specificity, and construct boundaries on the model itself \citep{kojima2022large}. The \emph{expanded} zero-shot condition used in the present simulation partially relaxes this constraint by supplying a construct definition and a small set of item-format requirements, thereby activating the context and output-format components of the prompt structure while still withholding examples. This design allows for a graded test of how much information, short of exemplars, is needed to improve output quality.

Few-shot prompting extends this logic by embedding a small set of examples directly into the prompt (e.g., sample items of the desired style and psychometric quality), which anchors outputs to a target register and implicitly communicates the expected format \citep{brown2020language}. The technique draws on the same in-context learning mechanism that underlies much of modern LLM behavior: exposure to even a handful of demonstrations can meaningfully improve task performance without any modification to model parameters \citep{chen2025,liu2023prompting}. In prompting research, these examples function as what might loosely be called embedded training signal, but not in the traditional sense of a large curated corpus: as a compact set of positive examples that constrain the space of possible outputs. Research on one-shot versus few-shot designs suggests that a single example may suffice for simpler tasks or highly capable models, while multiple examples provide additional guidance for more complex tasks or less capable ones \citep{brown2020language}. This technique is closely related to ``prompt programming'' approaches that treat prompts as structured inputs capable of controlling and stabilizing model behavior \citep{reynolds2021prompt}.

Persona prompting introduces a role or identity frame (e.g., ``You are an expert psychometrician developing a personality assessment'') to bias the model toward domain-consistent language, decision rules, and tone \citep{liu2024evaluatinglargelanguagemodel}. Conceptually, persona prompting functions as a high-level behavioral prior that shapes what the model attends to and how it interprets task demands \citep{olea2024evaluating,touvron2023llama}. Assigning a role to the model has been shown to improve accuracy on some knowledge-intensive benchmarks relative to role-free prompts \citep{kong2024better}, though the benefits are inconsistent and depend on factors such as the persona chosen and its alignment with the task domain \citep{zheng2024helpful}. This technique is widely used in applied settings, including psychological personality assessment \citep{de2023improved,jiang2024personallm}, because it tends to increase coherence and domain alignment, even when the requested output remains purely textual.

The most sophisticated technique in the present simulation is adaptive prompting. \citet{spliethover2025} introduce adaptive prompting as a method that selects the most effective prompt composition for each input instance from a predefined pool, using a learned predictor to identify the best configuration for that specific example. The present simulation adopts a related but operationally distinct form of adaptivity, grounded in the iterative refinement paradigm \citep{lightman2023letsverifystepstep}: the model is dynamically provided with its own previously generated items and explicitly instructed to avoid duplication (e.g., ``Do NOT repeat or rephrase anything in this list\ldots''). This design instantiates a generate-evaluate-revise cycle analogous to the prompt chaining logic described in the broader prompt engineering literature \citep{chen2025,russell2024generative}, in which each generation step is conditioned on the outputs of prior steps and targeted feedback. The approach is especially relevant to large-scale item generation because LLMs often exhibit degenerative repetition, where later outputs converge toward semantically similar phrasing in the absence of external constraints. A growing literature on iterative refinement, including self-reflection, self-revision, and feedback-conditioned prompting, demonstrates that repeatedly conditioning the model on its prior outputs and explicit corrective signals can substantially improve uniqueness and reliability without any retraining \citep{krishna2024understandingeffectsiterativeprompting,sun-etal-2024-enhancing}.

\subsection{AI-GENIE}

While prompt engineering strategies can substantially influence LLM-generated text, their practical value for psychological measurement depends on how these strategies translate into high-quality items. More specifically, quality item pools should have little redundancy and high structural validity. The present study therefore embeds prompt engineering strategies within the \textbf{AI-GENIE} pipeline \citep{russell2024generative}, which provides a deterministic, network-based mechanism to evaluate these different prompting approaches and LLMs.

Before diving into the present simulation, we provide an overview of the AI-GENIE methodology which consists of these steps:

\begin{itemize}
\item
  Generate and Embed an Initial Item Pool
\item
  Perform an Initial Exploratory Graph Analysis (EGA) to Set a Baseline
\item
  Run Unique Variable Analysis (UVA) Iteratively to Reduce Redundancy
\item
  Run Bootstrap EGA (bootEGA) Iteratively for Structural Validation
\item
  Perform a Final EGA to Assess Reduction Quality
\end{itemize}

\subsubsection{Generate and Embed an Initial Item Pool}

The process begins by generating an initial set of candidate items, which is the focus of the current simulation. Next, AI-GENIE embeds each item using a language-model embedding system (specifically OpenAI's text-embedding-3-small (\citealp{OpenAIembeddings})), converting each item into a high-dimensional semantic vector (that is, embeddings ``translate'' human language into quantitative input for psychometric modeling). These embeddings allow the pipeline to operationalize relationships among items based on meaning similarity, enabling the construction of a network.

\subsubsection{Perform an Initial Exploratory Graph Analysis (EGA) to Set a Baseline}

With the embedded item network in hand, the pipeline performs an initial Exploratory Graph Analysis (EGA; \citealp{GolinoEpskamp2017EGA}) to estimate the pool's dimensional structure as a baseline before any reduction steps. EGA is a network psychometric method for estimating the dimensional structure of a set of variables by modeling them as a network (as opposed to a traditional factor model). It first estimates a network (e.g., using Triangulated Maximally Filtered Graph (TMFG; \citealp{massara2016network}) and Extended Bayesian Information Criterion Glasso (EBICglasso; \citealp{foygel2010extended}; \citealp{friedman2008sparse})) where nodes are items and edges represent statistical or semantic relationships among them. Then, it applies a community detection algorithm (such as Walktrap (\citealp{Pons2005ComputingCC})) to identify clusters of densely connected nodes, which are interpreted as latent dimensions or factors. Because the clusters emerge from patterns of connectivity, EGA provides an empirically driven way to discover how many dimensions exist and which items belong to each dimension. The resulting dimensional solution can be evaluated using Normalized Mutual Information (NMI; \citealp{nmi}). NMI is a metric for quantifying how similar two different clusterings of the same set of items are, with scores closer to 0 indicating low accuracy, or total dissimilarity, and scores closer to 1 indicating perfect accuracy. In this context, it's used to evaluate how well the communities recovered by EGA match the known, ``true'' item groupings in simulation (e.g., whether items intended to measure Neuroticism actually cluster together).

\subsubsection{Run Unique Variable Analysis (UVA) Iteratively to Reduce Redundancy}

AI-GENIE then reduces the pool in two iterative phases designed to remove redundant and unstable items. First, Unique Variable Analysis (UVA; \citealp{christensen2023unique}) detects redundancy using weighted topological overlap (wTO; \citealp{zhang2005general}) within a Gaussian graphical model framework. In a network view, redundant items will look like they share nearly the same pattern of connections (neighbors and edge strengths). That is, two redundant items will have high wTO if they connect to the same other items with similar edge strengths. UVA identifies sets of items that exceed a redundancy cutoff (a wTO threshold). From each redundant pair or set, only the item with lowest overall overlap with the rest of the pool (i.e., the most unique representative of that set) is retained. The remaining items in that redundant cluster are removed because they contribute little unique coverage. Redundant items are removed iteratively until no further redundant pairs or sets remain.

\subsubsection{Run Bootstrap EGA (bootEGA) Iteratively for Structural Validation}

After redundancy removal, the pipeline uses bootstrap EGA (bootEGA; \citealp{psych3030032}) to assess structural stability. BootEGA is an extension of EGA that evaluates the stability and replicability of an estimated dimensional structure. BootEGA then tracks how consistently each item is assigned to the same dimension across resamples. Items that frequently ``wander'' between dimensions accross the bootstrap (low assignment consistency) are interpreted as poor quality items that destabilize the network structure, while items that remain in the same cluster are considered stronger and more reliable. Items that fall below a stability threshold are removed from the item pool. Then, this process is repeated until all remaining items show high stability.

\subsubsection{Perform a Final EGA to Assess Reduction Quality}

In the final step of the AI-GENIE pipeline, an EGA network is computed on the reduced item pool so the NMI can be recalculated and compared to the baseline calculated pre-reduction. This step is a post-pruning structural check to confirm that the reduced item pool recovers the intended dimensions better than the original pool. That is, after a successful AI-GENIE reduction, the NMI is higher than the initial NMI, indicating that the pruning removed less stable or highly redundant items while improving the recovery of the target structure.

\section{Methodology}

The present study used a Monte Carlo simulation to evaluate how prompting strategies influence the psychometric quality of AI-generated personality assessment items within the AI-GENIE framework. Each unique combination of experimental conditions was replicated 100 times, yielding a large set of independent item pools and reduction outcomes. Each replicate had at least 60 AI-authored items.

Experimental factors included:

\begin{itemize}
\item
  Four language models (GPT-4o, GPT-5.1, GPT-OSS-20B, and GPT-OSS-120B)
\item
  Three LLM model temperatures (0.5, 1, 1.5)
\item
  Five personality traits corresponding to \citeauthor{john1999bigfive}'s \citeyear{john1999bigfive} Big Five model (Openness, Conscientiousness, Extraversion, Agreeableness, Neuroticism). These are the OCEAN personality traits.
\item
  Six prompt designs (Basic, Expanded, Few Shot, Persona, Persona + Few Shot, Persona + Few Shot + Adaptive).
\item
  Two EGA network estimation models\footnote{For every generated item pool within each replication, the AI-GENIE reduction pipeline was executed under two alternative EGA model specifications. These EGA conditions did not involve regenerating new items; rather, the same generated item pool was embedded once and then passed through AI-GENIE under each EGA model to evaluate whether downstream structural validation and reduction results differed as a function of the network estimation method.} (TMFG, EBICglasso)
\end{itemize}

The first experimental factor is the LLM model used to generate the items. Two of the models are open-source (GPT-OSS 120B and 20B (\citealp{OSS})), and two of the models are proprietary (GPT-4o (\citealp{openaiGPT4o}) and GPT-5.1 \citep{gpt5}). The second experimental factor is the model's temperature setting. Temperature is a hyperparameter of an LLM that, very generally, sets the level of creativity one should expect from the output (see exception; \citet{peeperkorn2024temperature}). By increasing the temperature, an LLM would be more likely to select atypical tokens, inducing a unique response. Conversely, lower temperatures produce more predicable responses.

The third experimental factor is the OCEAN personality trait. Therefore, we will be creating item pools that target each of the Big Five \citep{john1999bigfive} personality traits. Th Big Five model was selected because we used this model in our simulation that demonstrates the efficacy of AI-GENIE \citep{russell2024generative}. Additionally, relying on the Big Five is deliberate because it is heavily represented in psychological research (and likely LLM training data), so it functions as a rigorous first test of whether AI-GENIE can match expert-developed measures under favorable, well-understood conditions. Once a pattern is observed, further research can test the robustness of the findings in the context of less established constructs, where training-data coverage and theory are weaker.

Several attributes associated with the targeted OCEAN personality traits were provided to the model. Including these trait attributes helps ensure that the items generated were not unidimensional and that they targeted several important aspects of the given trait. The attributes for each trait were as follows:

\begin{enumerate}
\item
  \emph{Openness}: creative, perceptual, curious, and philosophical
\item
  \emph{Conscientiousness}: organized, responsible, disciplined, and prudent
\item
  \emph{Neuroticism}: anxious, depressed, insecure, and emotional
\item
  \emph{Agreeableness}: cooperative, compassionate, trustworthy, and humble
\item
  \emph{Extraversion}: friendly, positive, assertive, and energetic
\end{enumerate}

The fourth experimental factor is the prompt design used to generate the items. The \emph{basic} prompt was a zero-shot design that plainly instructed the model to generate items that targeted each of the provided trait attributes. For example, the basic prompt used to generate `extraversion' items was as follows:

\begin{quote}
Generate novel items that assess the personality trait 'extraversion' from the 'Big Five' personality model. Extraversion has the following attributes of interest: friendly, positive, assertive, and energetic. For EACH attribute, write EXACTLY TWO (2) single-sentence, first-person items each reflecting ONLY that given attribute. Do NOT add any attributes or leave any out."
\end{quote}

At this point, the reader may be wondering why we would ask the model to produce only two items per attribute when a target of 60 total items is desired. When using an Application Programming Interface (API) to access these models, there are token limitations which prevent the model from outputting all 60 items in a single API call. Therefore, to generate 60 items, the prompt had to be parsed many times per sample.

The \emph{expanded} prompt elaborated on the instructions a bit further:

\begin{quote}
Ensure that each item is extremely high-quality, psychometrically robust, and concise. Each item should be novel, so be creative; aim for breadth across these attributes. Items should be polished and ready for immediate practical use.
\end{quote}

The \emph{few shot} prompt further appended a list of examples for the model to emualte in terms of quality and structure. These items were pulled from \citeauthor{john1999bigfive}'s \citeyear{john1999bigfive} assessment verbatim:

\begin{itemize}
\item
  I am someone who is compassionate, has a soft heart.
\item
  I am someone who starts arguments with others.
\item
  I am someone who is dependable, steady.
\item
  I am someone who has difficulty getting started on tasks.
\item
  I am someone who has an assertive personality.
\item
  I am someone who rarely feels excited or eager.
\item
  I am someone who is moody, has up and down mood swings.
\item
  I am someone who stays optimistic after experiencing a setback.
\item
  I am someone who is inventive, finds clever ways to do things.
\item
  I am someone who avoids intellectual, philosophical discussions.
\end{itemize}

To prevent the model from recycling the content of these examples, the model was also instructed to refrain from regurgitating \citeauthor{john1999bigfive}'s \citeyear{john1999bigfive} ideas:

\begin{quote}
Here are some EXAMPLE items that you must emulate in terms of QUALITY and item STRUCTURE only, but do NOT reuse any of these examples' content. The content of the items you generate must be entirely unique.
\end{quote}

The \emph{persona} prompt added an appropriate system role to the \emph{expanded} prompt. A system role is a secondary prompt that primes the model with an expert persona to improve its output. For example, our system role for the extraversion trait was as follows:

\begin{quote}
You are an expert psychometrician and test developer specializing in personality assessment. Your task is to create high-quality, psychometrically robust items for a personality inventory measuring 'extraversion' from the 'Big Five' model of personality.
\end{quote}

The \emph{persona + few shot} prompting condition simply added this system role to the \emph{few shot} prompt. Likewise, the \emph{persona + few shot + adaptive} prompt approach added the \emph{adaptive} component. Typically, we ran the prompt 8-15 times to build sample of at least 60 items total. The adaptive component included a running list of the items that the model had already generated in all of these previous outputs. For example, this adaptive component would tell the model the following:

\begin{quote}
Do NOT repeat, rephrase, or reuse the content of ANY items from this list of items you've already generated for 'extraversion': 
\end{quote}

\begin{enumerate}
\item
  \emph{previously generated extraversion item \#1.}
\item
  \emph{previously generated extraversion item \#2.}
\item
  \emph{previously generated extraversion item \#3.} and so on\ldots
\end{enumerate}

See Table 1 for a visual breakdown of which prompt component was included in each of the six prompting conditions for this simulation.

Lastly, a basic instruction on the required formatting was appended to the end of every single prompt so the output could be adequately parsed. For extraversion items, for example, these instructions were as follows:

\begin{quote}
Return output STRICTLY as a JSON array of objects, each with keys attribute and statement, e.g.:
[{"attribute":"friendly", "statement": "Your item here."}, ...]
This JSON formatting is EXTREMELY important. Do NOT include any explanations, commentary, or markdown. Output only the JSON. The "attribute" key should ONLY have these EXACT values: friendly, positive, assertive, and energetic.
\end{quote}

\begin{table}[ht]
\centering
\caption{\label{tab:prompt_components} Components Included in Each Prompting Design}
\resizebox{\textwidth}{!}{%
\begin{threeparttable}
\begin{tabular}{lcccccc}
\toprule
 & \textbf{BAS} & \textbf{EXP} & \textbf{FS} & \textbf{PER} & \textbf{PER+FS} & \textbf{PER+FS+A} \\
\midrule
Essential Instructions  & \checkmark & \checkmark & \checkmark & \checkmark & \checkmark & \checkmark \\
More Detailed Instructions       & --    & \checkmark & \checkmark & \checkmark & \checkmark & \checkmark \\
A system role prompt  & --    & --    & --    & \checkmark & --    & \checkmark \\
A list of item examples                     & --    & --    & \checkmark & --    & \checkmark & \checkmark \\
A list of items generated thus far & --    & --    & --    & --    & --    & \checkmark \\
\bottomrule
\end{tabular}
\begin{tablenotes}
\small
\item \textit{Note.} The prompt components used to construct the final prompt given to the model for each condition. The basic (BAS) prompt condition only received minimal instructions, whereas the expanded (EXP) prompt condition received more detailed instructions in addition to the minimal instructions. The other conditions built upon the expanded prompt condition further by adding components like a model persona and item examples.
\end{tablenotes}
\end{threeparttable}%
}
\end{table}

\section{Results}

\begin{figure}

{\centering \includegraphics[width=0.98\linewidth]{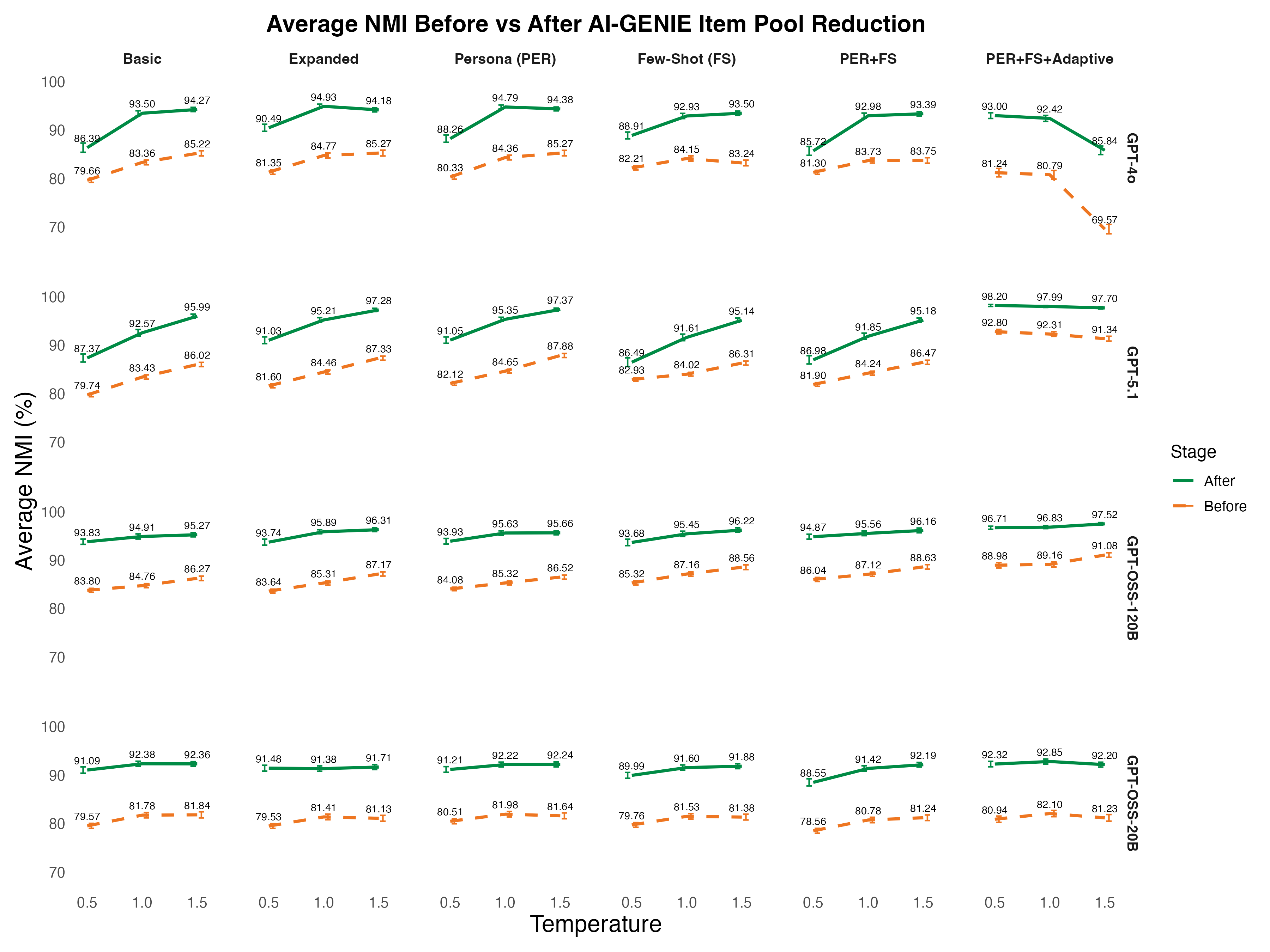} 

}

\caption{Average NMI before and after AI-GENIE reduction across prompting conditions, models, and temperatures. AI-GENIE reliably improved NMI across virtually all conditions. Adaptive prompting substantially raised the pre-reduction NMI floor for the newest models, with GPT-4o at temperature 1.5 as the sole condition where post-reduction NMI fell below the non-adaptive baseline.\label{fig:befafter}}
\end{figure}

Figure \ref{fig:befafter} provides an overview of average NMI before and after AI-GENIE reduction across all models, temperatures, and prompting conditions. Several patterns are immediately apparent: AI-GENIE reliably improves NMI across virtually all conditions (the blue post-reduction line consistently exceeds the orange pre-reduction line); adaptive prompting substantially raises the pre-reduction NMI floor for the newest models; and GPT-4o at temperature 1.5 under adaptive prompting is the sole condition where post-reduction NMI falls below the non-adaptive baseline, motivating closer examination of that model's behavior.

\begin{figure}

{\centering \includegraphics[width=0.95\linewidth]{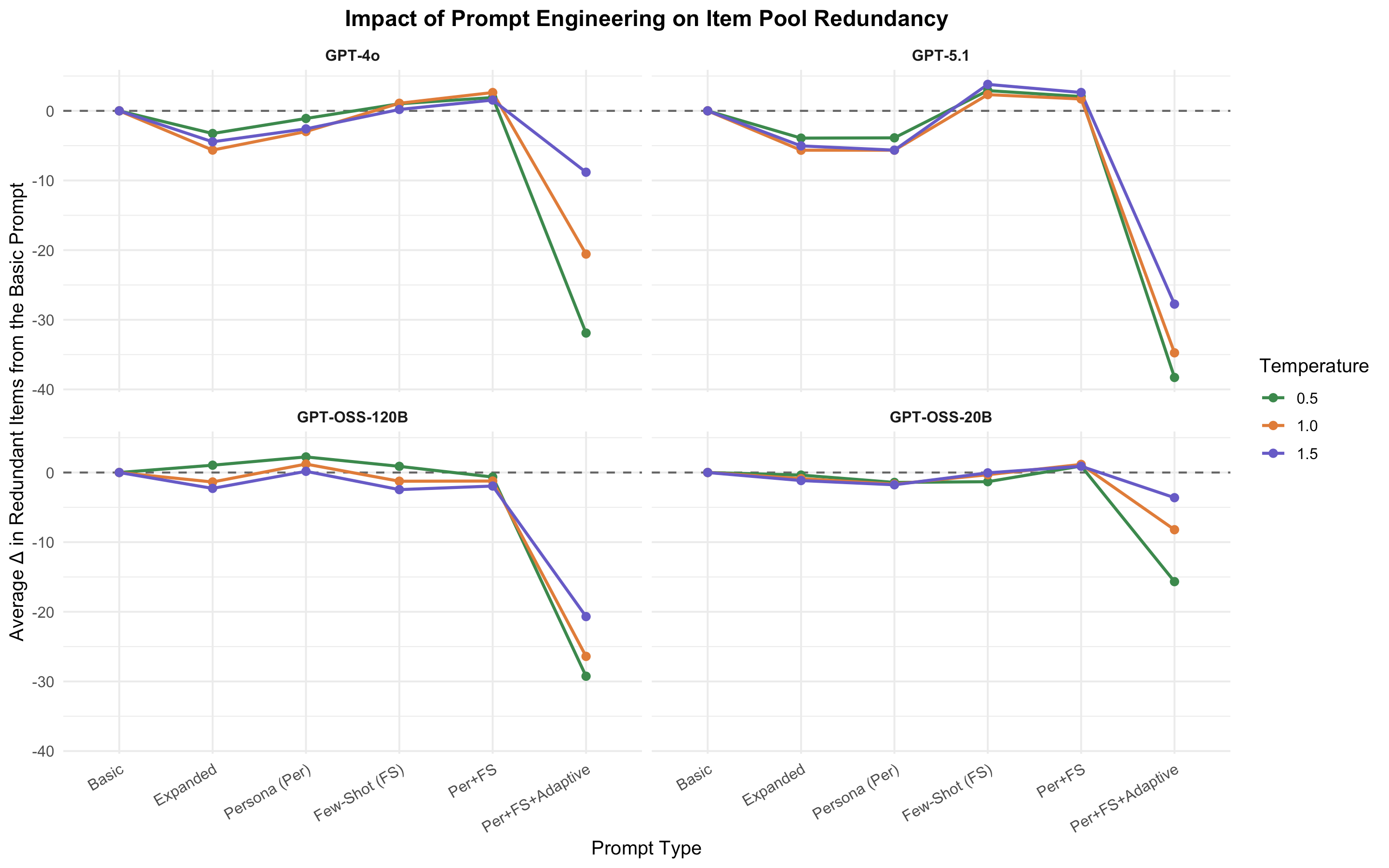} 

}

\caption{The average number of items removed during the UVA step of AI-GENIE relative to the basic prompt condition. The redundancy reduction is most notable for the adaptive prompting condition, which shows substantial gains over the baseline.\label{fig:UVA-remove}}
\end{figure}

The adaptive prompting strategy, especially when paired with newer models, proved to be especially powerful. Adaptive prompting produced large, systematic improvements in the quality of item pools across multiple outcome measures. However, the magnitude of improvement over the basic prompt baseline depended strongly on model capability and temperature sensitivity.

Most strikingly, the adaptive condition drastically reduced initial item pool redundancy by roughly 93.7\%. We observed similarly large proportional decreases of of 88.5\%, 83.3\%, and 53.4\%, for GPT-4o, GPT-OSS-120B, and GPT-OSS-20B respectively. For almost all experimental conditions, these reductions in redundancy were accompanied by gains in the final structural validity. Relative to the basic prompt, the final NMI increased by an average of 10.8\%, 5.4\%, and 1.7\% for GPT-5.1 at temperatures 0.5, 1.0, and 1.5, respectively, yielding near-ceiling final NMI values. Attenuated baseline improvements were observed for GPT-OSS-120B (1.9-2.9\% across temperatures), and GPT-OSS-20B showed modest gains (0-1.2\%). Adaptive prompting also substantially increased the number of items retained after AI-GENIE filtering, particularly for the newest models (e.g., GPT-5.1 had an average final sample size of 56--57 items under adaptive prompting and only 16--29 items under the basic prompt). GPT-4o represents a notable exception to the otherwise consistent pattern of adaptive gains. Although adaptive prompting reduced UVA removals for GPT-4o, improvements in the final NMI were not monotonic across temperatures. At the highest temperature, adaptive prompting resulted in an 8.5\% decrease relative to the basic prompt. This finding suggests an unusual, model-specific sensitivity to adaptive prompting at higher temperatures.

\subsection{Redundancy Analysis Using UVA}

We first examined how prompt design influenced redundancy in the initial item pools, operationalized as the number of items removed during the UVA step of AI-GENIE.

Across all models and temperatures, the adaptive prompting condition produced substantially fewer UVA removals than the basic prompt and all other non-adaptive prompt designs (Figure \ref{fig:UVA-remove}). This effect was especially pronounced for the newest LLM, GPT-5.1. Averaged across temperature settings, GPT-5.1 removed approximately 35.9 items under the basic prompt, compared to only 2.3 items under adaptive prompting, an absolute reduction of 33.6 items on average, corresponding to a 93.7\% decrease in redundancy. Importantly, this reduction was robust across temperature levels, with redundancy removals remaining a small handful of items.

A similar qualitative pattern was observed for GPT-4o and GPT-OSS-120B. GPT-4o exhibited an average reduction from 23.1 items removed under the basic prompt to 2.7 items under adaptive prompting (an 88.5\% reduction), while GPT-OSS-120B decreased from 30.6 to 5.1 items (an 83.3\% reduction). The smaller GPT-OSS-20B model showed a more modest benefit, with adaptive prompting reducing average UVA removals from 17.1 to 8.0 items (a 53.4\% reduction).

Non-adaptive prompting strategies, \emph{Expanded}, \emph{Few-Shot}, \emph{Persona}, and \emph{Persona + Few-Shot}, produced comparatively small and inconsistent changes relative to the basic prompt. In several cases, these non-adaptive designs yielded equal or greater redundancy than the basic condition, underscoring that redundancy reduction was not a general consequence of adding prompt complexity, but rather a distinctive effect of the adaptive component specifically.

\begin{figure}

{\centering \includegraphics[width=0.98\linewidth]{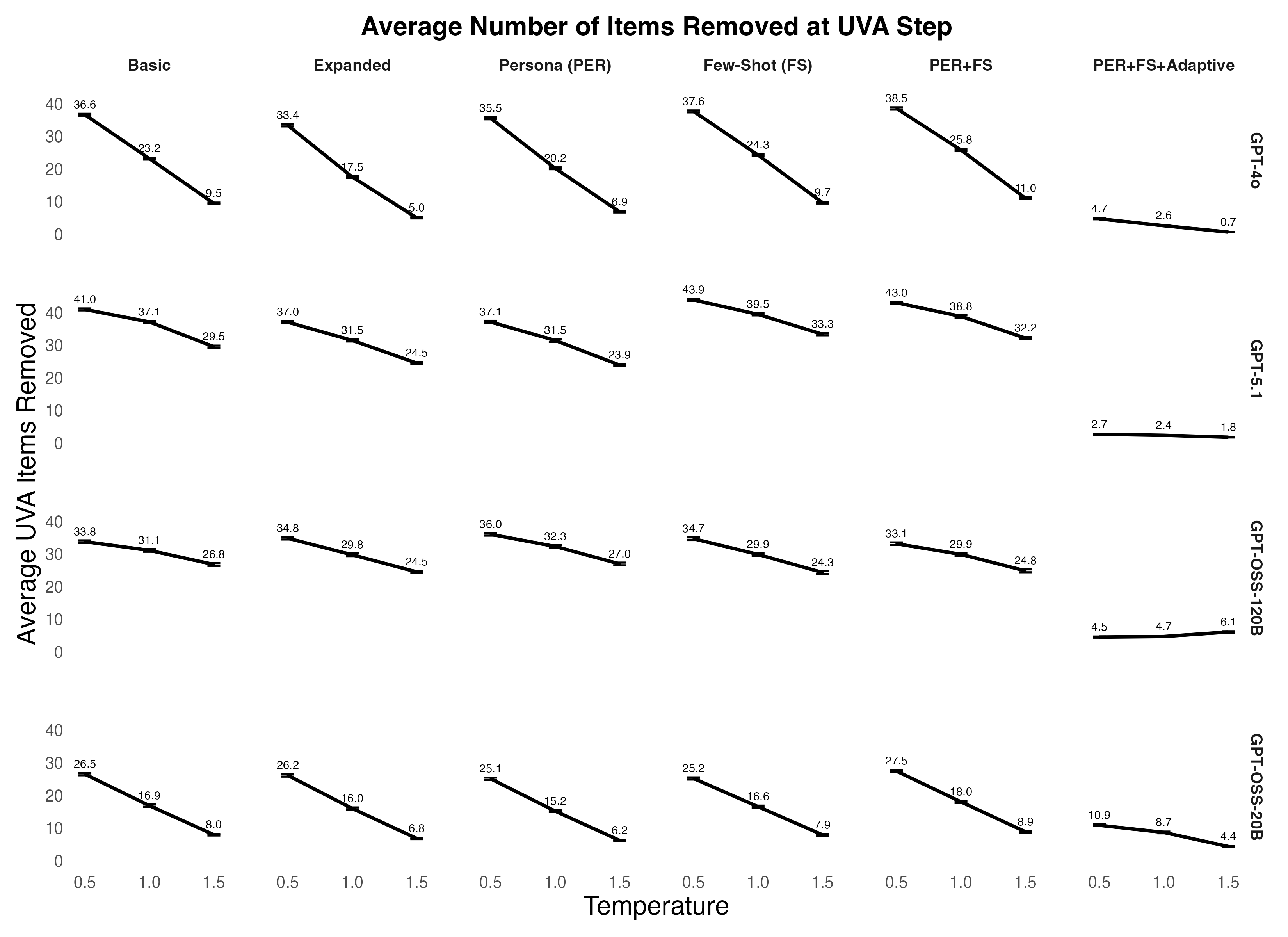} 

}

\caption{Average number of items removed at the UVA redundancy step across prompting conditions, models, and temperatures. Adaptive prompting (PER+FS+Adaptive) produced dramatically fewer removals than all other conditions, with reductions most pronounced for the newest models.\label{fig:UVA}}
\end{figure}

The absolute scale of this redundancy is notable: under the basic prompt at the lowest temperature, GPT-5.1 flagged an average of 41.0 items per pool as redundant (approximately 68\% of all generated items) underscoring just how severe degenerative repetition can be in the absence of adaptive constraints (Figure \ref{fig:UVA}).

\subsection{Pre-Reduction Quality Check Using NMI}

\begin{figure}

{\centering \includegraphics[width=0.98\linewidth]{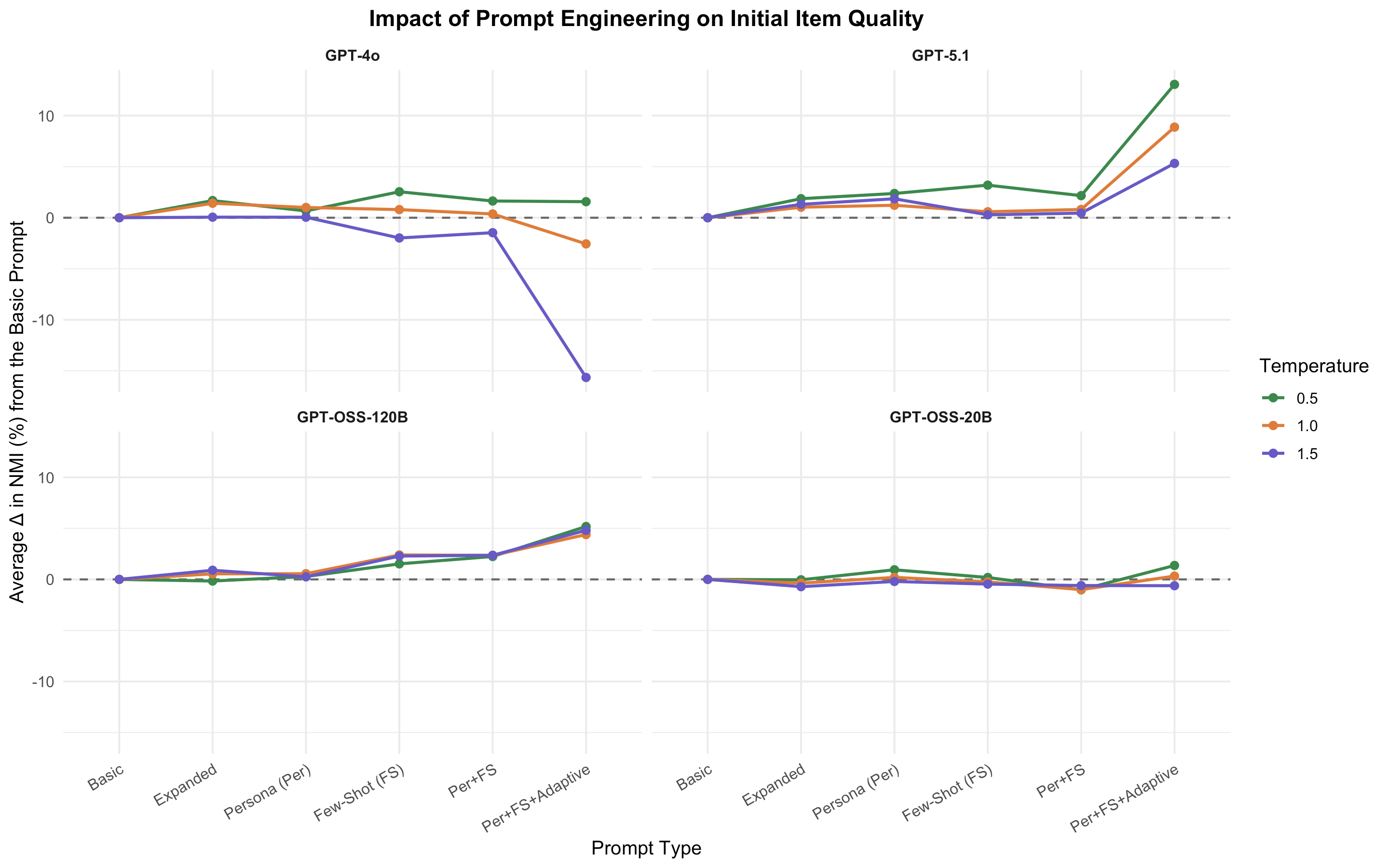} 

}

\caption{The average NMI before AI-GENIE reduction relative to the basic prompt condition. Adaptive prompting produced improvements in accuracy over the baseline for GPT-5.1 and GPT-OSS-120B. On the other hand, GPT-OSS-20B shows very modest gains while GPT-4o's high and default temperature models show a dip in the initial NMI.\label{fig:initial}}
\end{figure}

Consistent with the redundancy results, adaptive prompting produced the largest improvements in initial NMI for the newest models (Figure \ref{fig:initial}). Recall that the initial NMI (or accuracy) quantifies how well the EGA model recovered the known communities (i.e., Big Five Personality traits) before any pruning is done.

For GPT-5.1, adaptive prompting increased the initial NMI relative to the basic prompt by 13.1\%, 8.9\%, and 5.3\% at temperatures 0.5, 1.0, and 1.5, respectively, yielding initial NMI values exceeding 91\% across all temperatures. These gains were substantially larger than those observed for any non-adaptive prompting strategy, which typically improved initial NMI by only 1--3\% over the baseline. A similar but attenuated pattern was observed for GPT-OSS-120B, where adaptive prompting improved the initial NMI by approximately 4--5\% across temperatures. GPT-OSS-20B showed only minimal improvements in initial NMI under adaptive prompting (generally less than 1.5\%), indicating that the benefits of adaptive prompting scale with model capacity.

\subsection{Post-Reduction Quality Check Using NMI}

\begin{figure}

{\centering \includegraphics[width=0.98\linewidth]{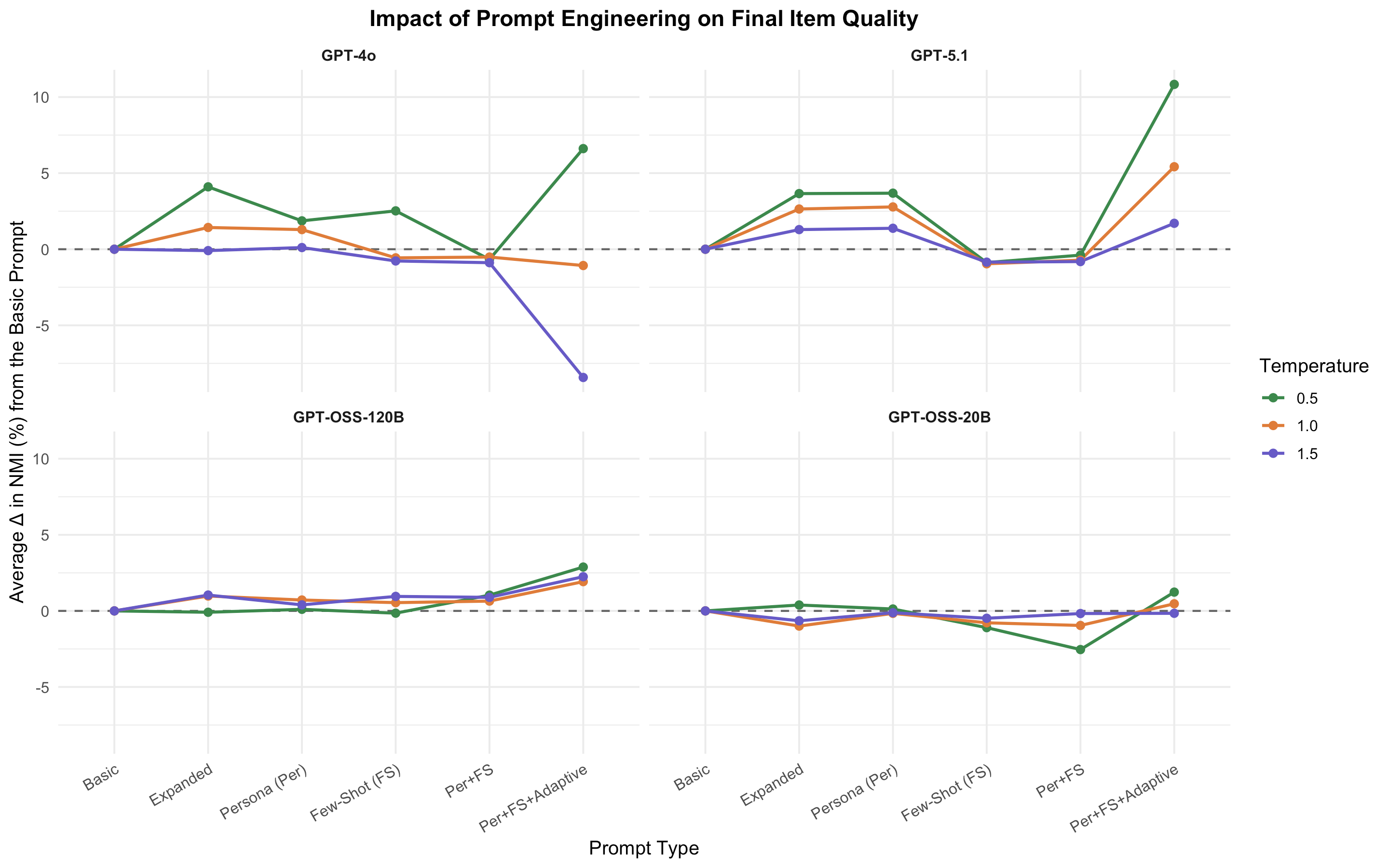} 

}

\caption{The average NMI after AI-GENIE implementation relative to the basic prompt condition. Adaptive prompting produced improvements in accuracy over the baseline for GPT-5.1, GPT-OSS-120B, and GPT-OSS-20B. However, GPT-4o's high temperature model provided a notable exception as adaptive prompting showed a dip in the final NMI.\label{fig:final}}
\end{figure}

As can be seen in Figure \ref{fig:final}, our results show that prompt design also influenced structural validity after AI-GENIE reduction. Whereas the initial NMI reflects the structure prior to AI-GENIE filtering, the final NMI reflects the combined effects of prompt design and the AI-GENIE pipeline, capturing how well the reduced item sets recover the intended dimensions.

Across models and temperatures, adaptive prompting produced the strongest post-reduction outcomes. This effect was most pronounced for GPT-5.1, which achieved almost perfect recovery under the adaptive condition. Relative to the basic prompt, adaptive prompting increased the final NMI by 10.8\%, 5.4\%, and 1.7\% at temperatures 0.5, 1.0, and 1.5, respectively, yielding average final NMI values of approximately 98\% across all temperatures. These gains substantially exceeded those observed for non-adaptive prompting strategies, which typically produced improvements of only 1--4\% over baseline.

GPT-OSS-120B exhibited a similar but attenuated pattern. Under adaptive prompting, final NMI increased by approximately 1.9\% to 2.9\% relative to the basic prompt across temperatures, resulting in final NMI values between 96\% and 97.5\% (Table \ref{tab:main_final_NMI_allDesigns}). While smaller in magnitude than the gains observed for GPT-5.1, these improvements were consistent across temperatures and indicate that adaptive prompting enhances post-reduction structural validity for large open-source models as well.

\begin{table}[ht]
\centering
\caption{\label{tab:main_final_NMI_allDesigns} Mean Final NMI After AI-GENIE Reduction}
\resizebox{\textwidth}{!}{%
\begin{threeparttable}
\begin{tabular}{lccccccc}
\toprule
 & \multicolumn{1}{c}{BASIC} & \multicolumn{1}{c}{EXPANDED} & \multicolumn{1}{c}{PERSONA} & \multicolumn{1}{c}{FEW SHOT} & \multicolumn{1}{c}{PER+FS} & \multicolumn{1}{c}{PER+FS+A} & \textbf{Grand Mean} \\
\textbf{gpt-4o} &  &  &  &  &  &  & \\
Temp 0.5 & 86.39 & 90.49 & 88.26 & 88.90 & 85.72 & 93.00 & \textbf{88.79} \\
Temp 1 & 93.49 & 94.93 & 94.79 & 92.93 & 92.98 & 92.42 & \textbf{93.59} \\
Temp 1.5 & 94.28 & 94.18 & 94.38 & 93.50 & 93.39 & 85.84 & \textbf{92.60} \\
\textbf{Mean} & \textbf{91.39} & \textbf{93.20} & \textbf{92.48} & \textbf{91.78} & \textbf{90.70} & \textbf{90.42} \\
\midrule
\textbf{gpt-5.1} &  &  &  &  &  &  & \\
Temp 0.5 & 87.37 & 91.01 & 91.06 & 86.50 & 87.01 & 98.20 & \textbf{90.19} \\
Temp 1 & 92.55 & 95.21 & 95.35 & 91.60 & 91.85 & 97.99 & \textbf{94.09} \\
Temp 1.5 & 95.99 & 97.28 & 97.37 & 95.14 & 95.18 & 97.70 & \textbf{96.44} \\
\textbf{Mean} & \textbf{91.97} & \textbf{94.50} & \textbf{94.59} & \textbf{91.08} & \textbf{91.35} & \textbf{97.96} \\
\midrule
\textbf{gpt-oss-120b} &  &  &  &  &  &  & \\
Temp 0.5 & 93.83 & 93.74 & 93.94 & 93.68 & 94.87 & 96.71 & \textbf{94.46} \\
Temp 1 & 94.91 & 95.89 & 95.63 & 95.45 & 95.55 & 96.83 & \textbf{95.71} \\
Temp 1.5 & 95.27 & 96.31 & 95.66 & 96.22 & 96.16 & 97.52 & \textbf{96.19} \\
\textbf{Mean} & \textbf{94.67} & \textbf{95.31} & \textbf{95.08} & \textbf{95.12} & \textbf{95.53} & \textbf{97.02} \\
\midrule
\textbf{gpt-oss-20b} &  &  &  &  &  &  & \\
Temp 0.5 & 91.08 & 91.48 & 91.22 & 89.99 & 88.55 & 92.32 & \textbf{90.77} \\
Temp 1 & 92.38 & 91.38 & 92.22 & 91.60 & 91.42 & 92.85 & \textbf{91.98} \\
Temp 1.5 & 92.36 & 91.71 & 92.24 & 91.88 & 92.19 & 92.20 & \textbf{92.10} \\
\textbf{Mean} & \textbf{91.94} & \textbf{91.52} & \textbf{91.89} & \textbf{91.15} & \textbf{90.72} & \textbf{92.46} \\
\midrule
\textbf{Grand Mean} & \textbf{92.49} & \textbf{93.63} & \textbf{93.51} & \textbf{92.28} & \textbf{92.07} & \textbf{94.47} \\
\bottomrule
\end{tabular}
\begin{tablenotes}
\item \textit{Note.} The average final NMI aftter AI-GENIE reduction. The highest final NMIs were recorded for the GPT-5.1 model under the adaptive prompting condition. However, the most modest NMIs were recorded for the same adaptive prompting condition when paired with the older GPT-4o model. 
\end{tablenotes}
\end{threeparttable}%
}
\end{table}

GPT-OSS-20B showed modest improvements in the final NMI under adaptive prompting, with gains generally below 1.5\% and average final NMI values near 92--93\%. This pattern mirrors the limited improvements observed in pre-reduction structure and redundancy reduction for this model, reinforcing the conclusion that the benefits of adaptive prompting scale with model capability.

GPT-4o again deviated from the general pattern. Although adaptive prompting improved final NMI at lower temperatures (e.g., 6.6\% at temperature 0.5), performance deteriorated sharply at the highest temperature. At temperature 1.5, adaptive prompting resulted in a final NMI of 85.8\%, representing an 8.4\% decrease relative to the basic prompt. This reversal contrasts with the stable or monotonic gains observed for the other models and underscores a model-specific sensitivity to adaptive constraints at higher levels of sampling stochasticity.

\subsection{AI-GENIE's Incremental Contribution Across Prompting Conditions}

\begin{figure}

{\centering \includegraphics[width=0.98\linewidth]{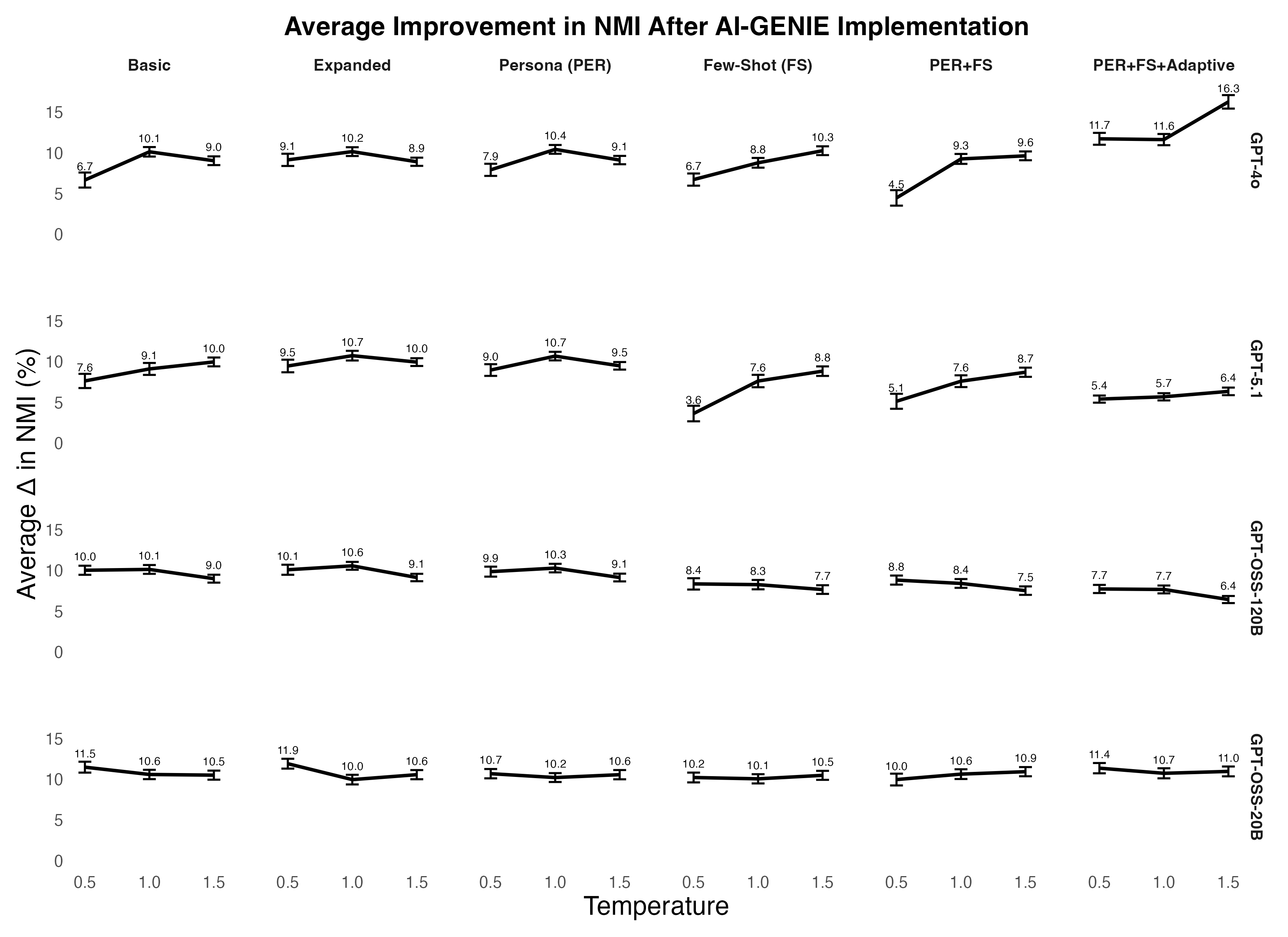} 

}

\caption{Average improvement in NMI delivered by AI-GENIE (pre- to post-reduction) across prompting conditions, models, and temperatures.\label{fig:incremental}}
\end{figure}

A complementary perspective on these results is provided by examining the magnitude of improvement delivered by AI-GENIE itself, that is, the gain from pre- to post-reduction NMI across prompting conditions (Figure \ref{fig:incremental}). For GPT-5.1 under non-adaptive conditions, AI-GENIE improved NMI by approximately 7.5-10\%, reflecting the pipeline's capacity to substantially refine lower-quality initial pools. Under adaptive prompting, this gain narrows to approximately 5-6\%. This apparent attenuation is not evidence of reduced pipeline effectiveness; rather, it reflects a ceiling dynamic. Adaptive prompting produces initial item pools with pre-reduction NMIs already exceeding 91\%, leaving less structural room for the pipeline to improve. This pattern confirms that adaptive prompting and AI-GENIE are complementary rather than redundant: the former elevates the quality of what enters the pipeline, while the latter provides deterministic refinement of whatever it receives. Notably, for GPT-OSS-20B, where adaptive prompting's effect on initial quality was minimal, AI-GENIE's incremental gain remains stable across all prompting conditions, consistent with the interpretation that the pipeline's contribution is largest when incoming quality is lowest.

\subsection{Item Pool Size after Reduction}

\begin{figure}

{\centering \includegraphics[width=0.98\linewidth]{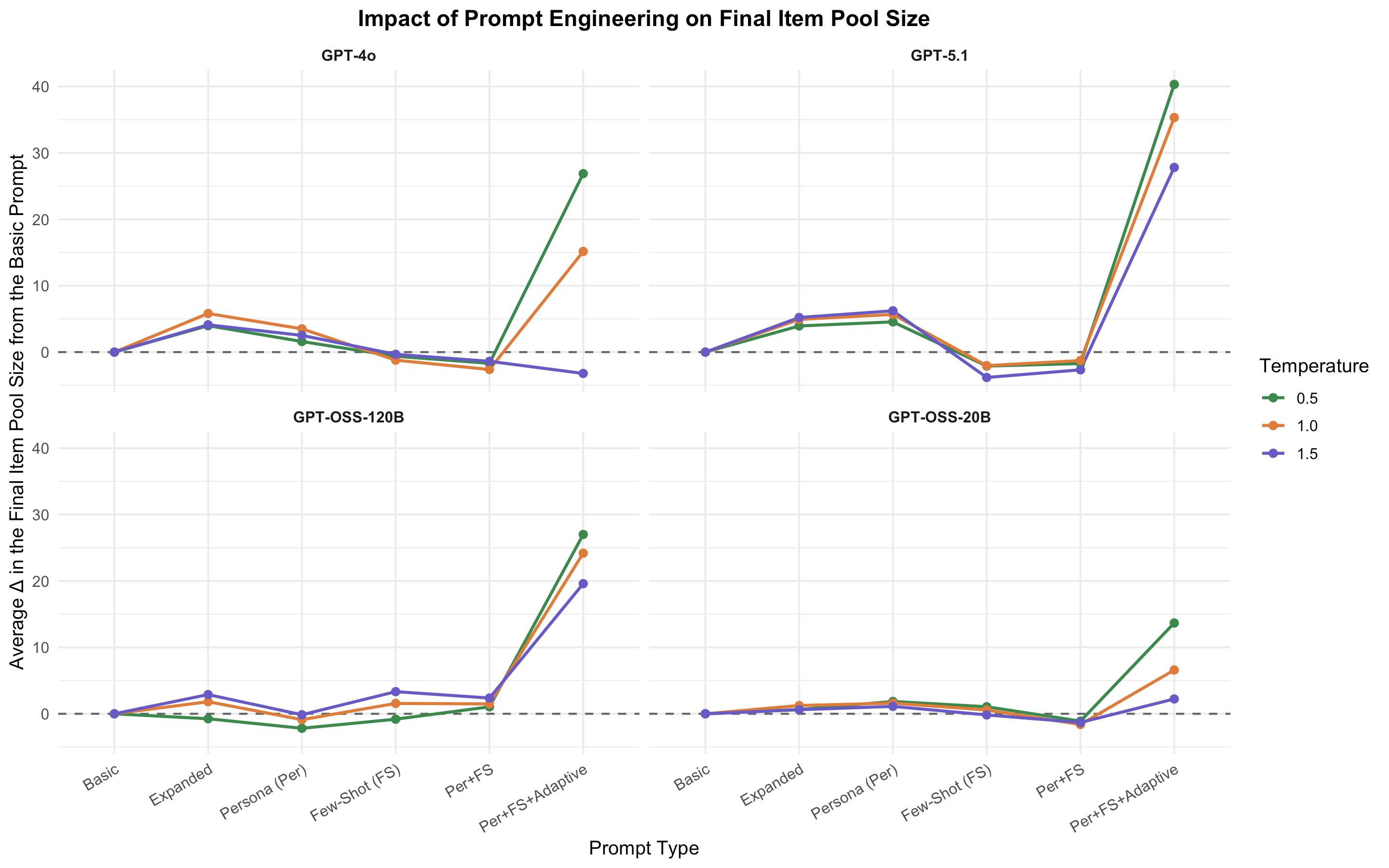} 

}

\caption{Average item pool size after AI-GENIE implementation relative to the basic prompt condition. Adaptive prompting retained the largest number of items for all conditions except GPT-4o's highest temperature model.\label{fig:finalsize}}
\end{figure}

Consistent with the preceding results, adaptive prompting yielded markedly higher item retention for the most capable models. For GPT-5.1, adaptive prompting resulted in an average of approximately 56--57 retained items across all temperature settings, compared to 16-29 items under the Basic prompt (Figure \ref{fig:finalsize}). This represents a more than twofold increase in retained items while simultaneously achieving near-ceiling post-reduction structural validity. A similar pattern was observed for GPT-OSS-120B, where adaptive prompting retained approximately 51-52 items, compared to only 25-32 items under the basic condition.

GPT-OSS-20B exhibited more modest retention gains under adaptive prompting, with the final item pool size increasing by approximately 6 to 14 items relative to the basic prompt, depending on temperature. This pattern mirrors the smaller improvements observed for redundancy reduction and final NMI in this model, suggesting that adaptive prompting yields diminishing returns for lower-capacity architectures.

For GPT-4o, adaptive prompting increased item retention at lower temperatures (e.g., final item pool size of roughly 46-49 items under adaptive prompting versus 20-33 items under basic prompting at temperatures 0.5 and 1.0), but this advantage diminished at the highest temperature. At temperature 1.5, adaptive prompting yielded slightly fewer retained items than the basic prompt, paralleling the decline in final NMI observed for this condition.

\subsection{Summary of Findings}

These findings suggest that adaptive prompting is the dominant driver of quality improvements in generative psychometric item development, with effects that scale strongly with model capability. For the newest model, GPT-5.1, adaptive prompting produced dramatic reductions in redundancy and large improvements in both pre and post-reduction community detection accuracy across all temperature settings, while simultaneously retaining substantially larger item pools after reduction. GPT-OSS-120B exhibited the same qualitative pattern with smaller but consistent gains, whereas GPT-OSS-20B showed more modest improvements. In contrast, non-adaptive prompting strategies yielded comparatively small, inconsistent, or negligible benefits relative to the basic prompt. GPT-4o proved to be somewhat of an exception to the otherwise impressive gains noticed when using adaptive prompting. Although adaptive prompting reduced redundancy for this model, its structural performance was highly sensitive to temperature, with marked declines in both initial and final NMI at the highest temperature. Overall, these findings indicate that adaptive prompting substantially enhances the efficiency and structural quality of AI-generated item pools when paired with sufficiently capable language models, motivating closer examination of model--prompt interactions and the mechanisms underlying these effects.

\section{Discussion}

The present study investigated how prompt engineering strategies interact with model capability to shape the quality of AI-generated assessment items within the AI-GENIE framework. Across a large Monte Carlo simulation, we found that adaptive prompting emerged as the dominant prompting approach, yielding dramatic reductions in redundancy and substantially greater item retention when paired with sufficiently capable LLMs. The older and smaller OSS model showed some benefits, and GPT-4o exhibited a distinctive pattern of temperature sensitivity that diverged from the general trend.

\subsection{Adaptive Prompting and ICL With Respect to Model Age and Size}

A central contribution of this work is demonstrating that the benefits of adaptive prompting scale sharply with model capability. For GPT-5.1, adaptive prompting reduced redundancy by over 90\%, produced near-perfect post-reduction NMIs, and retained substantially larger item pools. These gains were robust across temperature settings, indicating that the model was able to internalize and respect complex, cumulative constraints provided through adaptive prompting. By contrast, GPT-OSS-120B showed similar but smaller gains, and GPT-OSS-20B exhibited only modest improvements across outcomes.

This pattern aligns with the notion that larger and newer models are better able to treat prompts as a form of task-specific adaptation rather than as isolated instructions \citep{dong2024survey,wei2022emergent}. Empirical studies demonstrate that advanced prompting strategies often yield substantial gains only once models reach sufficient scale \citep{mehta2025scaling,wei2022chain}. Mechanistic and theoretical work further suggests that large transformer models can implement learning-like updates within their forward pass, enabling them to infer and apply abstract rules from context alone (\citealp{von2023transformers}; see also \citealp{liu2023pre}). From this perspective, adaptive prompting may function as a structured training signal, one that newer models can exploit more effectively because they possess the representational capacity to internalize set-level constraints (e.g., ``do not rephrase prior items'').

The most immediate effect of adaptive prompting was its impact on redundancy, as reflected by the dramatic reduction in items flagged by the UVA step of AI-GENIE. For GPT-5.1, adaptive prompting reduced UVA removals to less than 3 items on average across temperatures, indicating that the generated item pools were already diverse and almost entirely non-redundant before any pruning occurred. Importantly, non-adaptive prompting strategies did not consistently produce similar effects, and in some cases increased redundancy relative to the basic prompt.

Additionally, adaptive prompting substantially increased the number of usable items retained after reduction for the newest models. GPT-5.1 and GPT-OSS-120B retained roughly twice as many items under adaptive prompting as under the basic prompt, without sacrificing structural validity. Thus, researchers can generate larger, higher-quality initial pools that ultimately retain more items.

These findings underscore that prompt complexity alone is insufficient. What matters is whether the model can dynamically incorporate feedback about its prior outputs and adjust subsequent generations accordingly. Adaptive prompting directly targets the failure mode most relevant to large-scale item generation, semantic repetition, and appears to do so in a way that only sufficiently capable models can reliably exploit.

\subsection{Limitations}

A primary limitation of the present simulation is that it focuses exclusively on the Big Five personality traits. These constructs are unusually well-defined, broad, and widely represented in both psychological literature and general public discourse, which likely increases the probability that modern LLMs have learned rich semantic representations of them. As a result, the observed effectiveness of advanced prompting strategies may not generalize to constructs underrepresented in the literature. Although it should be noted that preliminary findings suggest that AI-GENIE works well for emerging constructs \citep{russell2024generative}.

Also, the present simulation is fully in silico and does not incorporate human expert evaluation. While this design is consistent with the goal of automating early-stage item development, human review remains a cornerstone of psychological measurement, particularly for ensuring that items represent the intended construct, avoid ambiguity or double-barreled wording, and adhere to ethical standards.

Additionally, these conclusions are anchored to a specific set of model families and versions. LLMs are not fixed entities. They are periodically updated through alignment changes, instruction tuning, and safety filters, and these updates can meaningfully alter performance characteristics such as repetition, creativity, specificity, and adherence to constraints. Consequently, newer models may shift the relative advantage of one prompting strategy over another, especially in tasks like item generation where subtle phrasing differences can have downstream consequences.

\subsection{Final Thoughts}

Prompt engineering can meaningfully shape the raw material of item pools, while AI-GENIE offers a deterministic, psychometrically grounded mechanism for transforming that material into structurally valid, nonredundant, and stable measures. Critically, the simulation shows that adaptive prompting is the standout methodological lever, simultaneously boosting structural validity to near-ceiling levels when used with the newest models and dampening the typical sensitivity to temperature settings. This simulation demonstrates a scalable path toward faster, more reproducible, and more cost-efficient measurement development.

AI-generated scale development is almost certainly on the horizon as a mainstream practice, but this paper makes a crucial contribution by showing that its rise does not have to resemble a methodological ``wild west.'' Rather than treating LLM-based item generation as an inherently unruly or unscientific shortcut, the work demonstrates how rigorous psychometric guardrails can be built directly into the workflow through structured, reproducible pipelines like AI-GENIE.

This precedent matters.

If AI is going to reshape measurement development, it can do so within a framework of transparency, replicability, and high standards, where prompt engineering becomes a disciplined methodological lever and psychometric validation remains the non-negotiable backbone. In that sense, the paper does not simply anticipate the future. It helps define the norms that should govern it.

\paragraph{Acknowledgements}
The authors did not preregister the study.

\paragraph{Competing Interests}
The authors report there are no competing interests to declare.

\paragraph{Author Contributions}
Lara L. Russell-Lasalandra: Conceptualization, Data Curation, Formal Analysis, Investigation, Methodology, Validation, Visualization, Writing -- Original Draft, Writing -- Review \& Editing; Hudson Golino: Conceptualization, Data Curation, Investigation, Methodology, Resources, Validation, Writing -- Original Draft, Writing -- Review \& Editing, Supervision.

\printbibliography

\end{document}